\documentclass[11pt,a4paper]{article}

\usepackage[hyperref]{acl2017}
\usepackage{times}
\usepackage{latexsym}
\usepackage{graphicx}
\usepackage{subfigure}
\usepackage{amssymb}
\usepackage{amsmath}
\usepackage{bm}
\usepackage{xspace}
\usepackage[english]{babel}
\usepackage{enumitem}

\usepackage{xcolor}

\usepackage{url}

\aclfinalcopy 

\newcommand{\DS}{\texttt{DS}\xspace}
\newcommand{\KB}{\texttt{KB}\xspace}

\newcommand{\RE}{\texttt{RE}\xspace}
\newcommand{\PR}{\texttt{PR}\xspace}
\newcommand{\TimeRE}{\textsc{TimeRE}\xspace}
\newcommand{\EntityRE}{\textsc{EntityRE}\xspace}
\newcommand{\NYT}{\texttt{NYT}\xspace}

\title{Learning with Noise: Enhance Distantly Supervised Relation Extraction with Dynamic Transition Matrix}

\author{Bingfeng Luo$^1$, Yansong Feng$^{*1}$, Zheng Wang$^2$, Zhanxing Zhu$^3$,
\\\textbf{Songfang Huang$^4$, Rui Yan$^1$ \and Dongyan Zhao$^1$}\\
  $^1$ICST, Peking University, China\\
  $^2$School of Computing and Communications, Lancaster University, UK \\
  $^3$Peking University, China\\
  $^4$IBM China Research Lab, China \\
  {\tt \{bf\_luo,fengyansong,zhanxing.zhu,ruiyan,zhaody\}@pku.edu.cn} \\
  {\tt z.wang@lancaster.ac.uk} \\
  {\tt huangsf@cn.ibm.com} \\}

\begin{document}
\maketitle

\begin{abstract}
Distant supervision significantly reduces human efforts in building
training data for many classification tasks. While promising,
this technique often introduces noise to the generated training data, which can
severely affect the model performance. In this paper, we take
a deep look at the application of distant supervision in relation extraction.
We show that the dynamic transition matrix can effectively characterize the noise in the training data built by distant supervision. 
The transition matrix can be effectively trained using a novel curriculum
learning based method without any direct supervision about the noise.
We thoroughly evaluate our approach under a wide range of extraction scenarios. 
Experimental results show that our approach consistently improves the extraction results and outperforms 
the state-of-the-art in various evaluation scenarios. 



\end{abstract} 
\section{Introduction}

Distant supervision (\DS) is rapidly emerging as a viable means for supporting various classification tasks -- from relation extraction~\cite{mintz2009distant} and sentiment classification~\cite{go2009twitter} to cross-lingual semantic
analysis~\cite{fang2016learning}.
By using knowledge learned from seed examples to label data, \DS automatically prepares large scale training data for these tasks.



While promising, \DS does not guarantee perfect results and often introduces noise to the
generated data. In the context of relation extraction, 
\DS works by considering sentences containing both the subject and object of a $<$\emph{subj}, \texttt{rel}, \emph{obj}$>$ triple as its supports. However, the generated data are not always perfect. For instance,  
\DS could match the knowledge base (\KB) triple, $<$\emph{Donald Trump},
\texttt{born-in}, \emph{New York}$>$  in \emph{false positive} contexts like \emph{Donald Trump worked in New York City}.
Prior works~\cite{takamatsu2012reducing,ritter2013modeling} show that \DS often mistakenly labels real positive instances as negative (\emph{false negative}) or
versa vice (\emph{false positive}), and there could be  confusions among positive labels as well. 
These noises can severely affect training and lead to poorly-performing models.

Tackling the noisy data problem of \DS is non-trivial, since there usually lacks of explicit supervision to capture the noise.
Previous works have tried to remove sentences containing unreliable syntactic patterns~\cite{takamatsu2012reducing}, 
design new models to capture certain types of noise or aggregate multiple predictions under the
\textit{at-least-one assumption}
that at least one of the aligned sentences supports the triple in \KB~\cite{riedel2010modeling,surdeanu2012multi,ritter2013modeling,min2013distant}.
These approaches represent a substantial leap forward towards making \DS more practical. however,  are either tightly couple to certain types of noise,
or have to rely on manual rules to filter noise, thus unable to scale.
%
Recent breakthrough in neural networks provides a new way to reduce the influence of incorrectly labeled data by aggregating multiple training instances attentively for relation classification, without explicitly characterizing the inherent noise~\cite{lin2016neural,zeng2015distant}.
Although promising, 
 modeling noise within neural network architectures is still in its early stage and much remains to be done.

In this paper, we aim to enhance \DS noise modeling by providing the capability to explicitly characterize 
the noise in the \DS-style training data within neural networks architectures.  We show that while noise is inevitable, it is possible to characterize the noise pattern  in a unified framework along with its original classification objective. Our key insight is that the \DS-style training  data typically contain useful clues about the noise pattern. For example, we can infer that since some people work in their birthplaces, \DS could wrongly label a training sentence describing a working place as a \texttt{born-in} relation.
Our novel approach to noisy modeling is to use a dynamically-generated transition matrix for each training instance to (1) characterize the possibility that the \DS labeled relation is confused and (2) indicate its noise pattern.  To tackle the challenge of no direct guidance over the noise pattern, we employ a curriculum learning based training method to gradually model the noise pattern over time, and utilize trace regularization to control the behavior of the transition matrix during training. Our approach is flexible -- while it does not make any assumptions about the data quality, the algorithm can make effective use
of the data-quality prior knowledge to guide the learning procedure when such clues are available. 

We apply our method to the relation extraction task and evaluate under various scenarios on two benchmark datasets. Experimental results show that our approach consistently improves both extraction settings, outperforming the state-of-the-art models in different settings. 

Our work offers an effective way for tackling the noisy data problem of \DS, making \DS more practical at scale. Our main contributions are to (1) design a \emph{dynamic} transition matrix structure to characterize the noise introduced by \DS, and (2) design a curriculum learning based framework to adaptively guide the training procedure to learn with noise.


\section{Problem Definition}
The task of distantly supervised relation extraction is to extract knowledge triples, $<$\emph{subj}, \texttt{rel}, \emph{obj}$>$, from free text with the training data constructed by aligning existing \KB triples with a large corpus.
Specifically, given a triple in \KB, \DS works by first retrieving all the sentences containing both \emph{subj} and \emph{obj} of the triple, and then constructing the training data by considering these sentences as support to the existence of the triple.
This task can be conducted in both the sentence and the bag levels.
The former  takes a sentence $s$ containing both $subj$ and $obj$ as input, and outputs the relation expressed
by the sentence between $subj$ and $obj$.
The latter setting
alleviates the noisy data problem by using the \textit{\textbf{at-least-one assumption}}
that at least one of the retrieved sentences containing both $subj$ and $obj$ supports the $<$\emph{subj}, \texttt{rel}, \emph{obj}$>$ triple.
It takes a bag of sentences $S$ as input where each sentence
$s\in S$ contains both $subj$ and
$obj$, and outputs  the relation between $subj$ and $obj$ expressed by this bag.

\begin{figure}[t!]
\begin{center}
\includegraphics[width=0.485\textwidth]{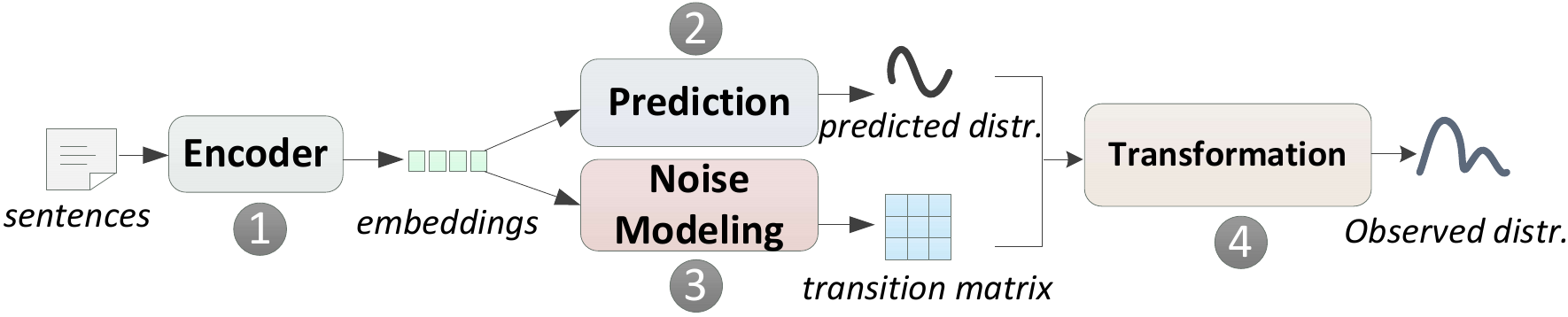}
\setlength{\belowcaptionskip}{-10pt}
\vspace{-5mm}	
\caption{Overview of our approach}
\vspace{-2mm}
\label{fig: denoise_framework}
\end{center}
\end{figure}

\section{Our approach}\label{sec:approach}

In order to deal with the noisy training data obtained through \DS, our approach follows four steps as depicted in Figure \ref{fig: denoise_framework}.
First, each input sentence is fed to a sentence encoder to generate an embedding vector. Our model then takes the sentence embeddings as input and produce a predicted relation
distribution, $\mathbf{p}$, for the input sentence (or the input sentence bag). At the same time, our model dynamically produces
a transition matrix, $\mathbf{T}$, which is used to characterize the noise pattern of  
sentence (or the bag). Finally, the predicted distribution is multiplied by the transition matrix to produce the observed relation
distribution, $\mathbf{o}$, which is used to match the noisy relation labels
assigned by \DS
while the predicted relation distribution $\mathbf{p}$ serves
as output of our model during testing.
One of the key challenges of our approach is on determining the element values of the transition matrix, which will be described in Section~\ref{sec:training}.

\subsection{Sentence-level Modeling}

\paragraph{Sentence Embedding and Prediction}
 In this work, we use a piecewise convolutional neural network~\cite{zeng2015distant} for sentence encoding, but other sentence embedding models can also be used.
We feed the sentence embedding to a full connection layer, and use \emph{softmax} to generate the predicted relation distribution, $\mathbf{p}$.

\paragraph{Noise Modeling}
%
First, each sentence embedding $\mathbf{x}$,
generated b sentence encoder, is passed to a full connection layer as a  non-linearity to obtain the
sentence embedding $\mathbf{x}_n$ used specifically for noise modeling.
We then use \emph{softmax} to calculate the transition matrix $\mathbf{T}$, for each sentence:
%
\begin{equation}\label{eq_tm}
T_{ij} = \frac{exp({\mathbf{w}_{ij}^T \mathbf{x}_n + b})}{\sum_{j=1}^{|\mathbb{C}|}{exp({\mathbf{w}_{ij}^T \mathbf{x}_n + b}})}
\end{equation}
where $T_{ij}$ is the conditional probability for the input sentence to be labeled as relation $j$ by \DS, given $i$ as the true relation, $b$ is a scalar bias,  $|\mathbb{C}|$ is the number of relations, $\mathbf{w}_{ij}$ is the weight vector characterizing the confusion between $i$ and~$j$. 

Here, we dynamically produce a transition matrix, $\mathbf{T}$, specifically for each sentence, but with the parameters ($\mathbf{w}_{ij}$) shared across the dataset. By doing so, we are able to adaptively characterize the noise pattern for each sentence, with a few parameters only. In contrast, one could also produce a global transition matrix for all sentences, with much less computation, where one need not to compute $\mathbf{T}$ on the fly (see Section \ref{sec:results_in_TimeRE}).

\paragraph{Observed Distribution}
When we characterize the noise in a sentence with a transition matrix $\mathbf{T}$,
if its true relation is $i$, we can assume that  $i$ might be erroneously labeled as relation $j$
by \DS  with probability $T_{ij}$.
We can therefore capture the observed relation distribution, $\mathbf{o}$, by
multiplying 
$\mathbf{T}$ and the predicted relation distribution, $\mathbf{p}$:
 \begin{equation}
\mathbf{o} = \mathbf{T}^T \bm\cdot \mathbf{p}
\label{eq_transition}
 \end{equation}
where 
$\mathbf{o}$ is then normalized to ensure $\sum_i{o_i}=1$.

Rather than using the predicted distribution $\mathbf{p}$ to directly match the relation labeled by \DS~\cite{zeng2015distant,lin2016neural},
here we utilize $\mathbf{o}$ to match the noisy labels during training and still use $\mathbf{p}$ as output during testing,
which actually captures the procedure of how the noisy label is produced and thus protects $\mathbf{p}$ from the noise.
%

\subsection{Bag Level Modeling \label{sec:baglevelmodeling}}
\paragraph{Bag Embedding and Prediction}
One of the key challenges for bag level model is how to aggregate the embeddings of individual sentences into the bag level.
In this work, we experiment two methods, namely average and attention aggregation~\cite{lin2016neural}.
The former calculates the bag embedding, $\mathbf{s}$, by averaging the embeddings of each sentence, and  then feed it to a \emph{softmax} classifier for relation classification.

The attention aggregation calculates an attention value, $a_{ij}$, for each sentence $i$ in the bag with respect to each relation $j$, and aggregates to  the bag level as  $\mathbf{s}_j$, by the following equations\footnote{While~\cite{lin2016neural} use bilinear function to calculate $a_{ij}$, we simply use dot product since we find these two functions perform similarly in our experiments.}:

\begin{equation}
\begin{aligned}
\mathbf{s}_j = \sum_i^{n}{a_{ij} \mathbf{x}_{i}}; \mbox{       }a_{ij} = \frac{exp(\mathbf{x}_i^T \mathbf{r}_j)}{\sum_{i'}^n{exp(\mathbf{x}_{i'}^T \mathbf{r}_j)}}
\end{aligned}
\label{att_sum}
\end{equation}
%
where $\mathbf{x}_{i}$ is the embedding of sentence $i$, $n$ the number of sentences in the bag, and $\mathbf{r}_j$ is the randomly initialized embedding for relation $j$.
In similar spirit to~\cite{lin2016neural},
the resulting bag embedding $\mathbf{s}_j$ is fed to a \emph{softmax} classifier 
to predict the probability of relation $j$ for the given bag.

\paragraph{Noise Modeling}
Since the transition matrix addresses the transition probability with respect to each true relation, the attention mechanism appears to be a natural fit for calculating the transition matrix in bag level.
Similar to attention aggregation above,
we calculate the bag embedding with respect to each relation using Equation~\ref{att_sum},  but with a separate set of relation embeddings $\mathbf{r'}_j$.
We then calculate the transition matrix, $\mathbf{T}$, by:
\begin{equation}
T_{ij} = \frac{exp({\mathbf{s}_i^T \mathbf{r'}_j  + b_i})}{\sum_{j=1}^{|\mathbb{C}|}{exp(\mathbf{s}_i^T \mathbf{r'}_j + b_i})}
\end{equation}
where $\mathbf{s}_i$ is the bag embedding regarding relation $i$, and $\mathbf{r'}_j$ is the embedding for relation $j$.

\section{Curriculum Learning based Training \label{sec:training}}

One of the key challenges of this work is  on how to train and produce the transition matrix to model the noise  in the training data without any direct guidance and human involvement.
A straightforward solution is to directly align the observed distribution, $\mathbf{o}$, with respect to the noisy labels by minimizing the sum of the two terms:
$CrossEntropy(\mathbf{o}) + Regularization$. However, doing so
does not guarantee that the prediction distribution, $\mathbf{p}$, will match the true relation distribution.
The problem is at the beginning of the training, we have no prior knowledge about the noise pattern, 
thus, both  $\mathbf{T}$ and $\mathbf{p}$ are less reliable, making the training procedure be likely to trap into some poor local optimum.
Therefore, we require a technique to guide our model to gradually adapt to the noisy training data, e.g., learning something simple first, and then trying to deal with noises.

Fortunately, this is exactly what curriculum learning can do.
The idea of curriculum learning~\cite{bengio2009curriculum} is simple: starting with the easiest aspect of a task, and leveling up the difficulty gradually,
which fits well to our problem.
We thus employ a curriculum
learning framework to guide our model to gradually learn how to characterize the noise. 
Another advantage is to avoid falling into poor local optimum.

With curriculum learning, our approach  provides the
flexibility to combine prior knowledge of noise, e.g., splitting a dataset into reliable and less reliable subsets,  to improve the
effectiveness of  the transition matrix and better model the noise.


\subsection{Trace Regularization}
Before proceeding to training details, we first discuss how we characterize the noise level of the data by controlling the trace of its transition matrix.
Intuitively, if the noise is small, the transition matrix $\mathbf{T}$ will tend to become an identity matrix, i.e., given a set of annotated training sentences,  the observed relations and their true relations are almost identical. Since each row of $\mathbf{T}$ sums to 1, the similarity between the transition matrix and the identity matrix can be represented by its trace, $trace (\mathbf{T})$. The larger the $trace(\mathbf{T})$ is, the larger the diagonal elements are, and the more similar the transition matrix $\mathbf{T}$ is to the identity matrix, indicating a lower level of noise. Therefore, we can characterize  the noise pattern by controlling the expected value of $trace(\mathbf{T})$ in the form of regularization. For example, we will expect a larger $trace (\mathbf{T})$ for reliable data, but  a smaller $trace(\mathbf{T})$  for less reliable data. Another advantage of employing trace regularization is that it could help reduce the model complexity  
and avoid overfitting.


\subsection{Training}
To tackle the challenge of no direct guidance over the noise patterns,
we implement a curriculum learning based training method to
first train the model without considerations for noise. In other words, we first focus on the loss from the prediction distribution $\mathbf{p}$ ,  and then take the noise modeling into account gradually along the training process, i.e., gradually increasing the importance of the loss from the observed distribution $\mathbf{o}$ while decreasing the importance of $\mathbf{p}$. In this way, the prediction branch is roughly trained before the model managing to characterize the noise, thus avoids being stuck into poor local optimum.
We thus design to minimize  the following loss function:
%
%
\begin{equation}
\begin{aligned}
L&=\sum_{i=1}^N{-((1-\alpha) log(o_{iy_{i}}) + \alpha log(p_{iy_{i}}))} \\
&- \beta trace(\mathbf{T}^{i})
\end{aligned}
\label{general_loss}
\end{equation}
where 0$<$$\alpha$$\le$1 and $\beta$$>$0 are two weighting parameters, $y_i$ is the relation assigned by \DS for the $i$-th  instance, $N$ the total number of training instances, $o_{iy_{i}}$ is the probability that the observed relation for the $i$-th instance is $y_i$, and $p_{iy_{i}}$ is the probability to predict relation $y_i$ for the $i$-th instance.

Initially, we set $\alpha$=1, and train our model  completely by minimizing the loss from the prediction distribution $\mathbf{p}$. That is, we do not expect to model the noise,  but focus  on the prediction branch at this time. As the training progresses, the prediction branch gradually learns the basic prediction ability. We then decrease $\alpha$ and  $\beta$ by 0$<$$\rho$$<$1 ($\alpha^*$=$\rho$$\alpha$ and $\beta^*$=$\rho$$\beta$) every $\tau$ epochs, i.e., learning more about the noise from the observed distribution $\mathbf{o}$ and allowing a relatively smaller $trace(\mathbf{T})$ to accommodate more noise.
The motivation behind is to put more and more effort on learning the noise pattern as the training proceeds, 
with the essence of curriculum learning.
This gradually learning paradigm significantly distinguishes from prior work on noise modeling for \DS seen to date. 
Moreover, as such a method does not rely on any extra assumptions,
it can serve as our default training method for $\mathbf{T}$.

\paragraph{With Prior Knowledge of Data Quality}
On the other hand, if we happen to have prior knowledge about which part of the training data is more reliable and which is less reliable, we can utilize this knowledge as guidance to design the curriculum.  
Specifically, we can build a curriculum by first training the prediction branch on the reliable data for several epochs, and then adding the less reliable data to train the full model. In this way, the prediction branch is roughly trained before exposed to more noisy data, thus is less likely to fall into poor local optimum.

Furthermore, we can take better control of the training procedure with trace regularization, e.g., encouraging larger $trace (\mathbf{T})$ for reliable subset and smaller $trace (\mathbf{T})$ for less relaibale ones.
Specifically, we propose to minimize:
\begin{equation}
\begin{aligned}
L=\sum_{m=1}^M{\sum_{i=1}^{N_m}{-log(o_{mi,y_{mi}})}} - \beta_m trace(\mathbf{T}^{mi})
\end{aligned}
\end{equation}
where $\beta_m$ is the regularization weight for the $m$-th data subset, $M$ is the total number of subsets, $N_m$ the number of instances in $m$-th subset, and  $\mathbf{T}^{mi}$, $y_{mi}$ and $o_{mi,y_{mi}}$ are the transition matrix, the relation labeled by \DS and the observed probability of this relation for the $i$-th training instance in the $m$-th subset, respectively. Note that different from Equation~\ref{general_loss}, this loss function does not need to initiate training by
minimizing the loss regarding the prediction distribution $\mathbf{p}$, since one can easily start by learning from the most reliable split first. 


We also use trace regularization for the most reliable subset, since there are still some noise annotations inevitably appearing in this split. 
Specifically, we expect its $trace(\mathbf{T})$ to be large (using a positive $\beta$) so that the elements of $\mathbf{T}$ will be centralized to the diagonal and $\mathbf{T}$ will be more similar to the identity matrix. As for the  less reliable subset, we expect the $trace (\mathbf{T})$ to be small (using a negative $\beta$) so that the elements of the transition matrix will be diffusive and $\mathbf{T}$  will be less similar to the identity matrix. In other words, the transition matrix is encouraged to characterize the noise.

Note that this loss function only works for sentence level models. For bag level models, since reliable and less reliable sentences are all aggregated into a sentence bag,  we can not determine which bag is reliable and which is not. However, bag level models can still build a curriculum by changing the content of a bag, e.g., keeping reliable sentences in the bag first, then gradually adding less reliable ones, and training with Equation~\ref{general_loss}, which could benefit from the prior knowledge of data quality as well.

\section{Evaluation Methodology}

Our experiments aim to answer two main questions:
(1) is it possible to model the noise in the training data generated through  \DS, even when there is no prior knowledge to guide us?
%
and (2) whether the prior knowledge of data quality can help our approach better handle the noise.

We apply our approach to both sentence level and bag level
extraction models, and evaluate in the situations where we do not have prior knowledge of
the data quality as well as 
where such prior knowledge is available.

\subsection{Datasets}
We evaluate our approach on two datasets.  
%

\paragraph{\TimeRE}
We build \TimeRE by
using \DS to align time-related Wikidata~\cite{vrandevcic2014wikidata} \KB triples to
Wikipedia text. It contains 278,141 sentences with 12
types of relations  between an entity mention and a time expression.
We choose to use time-related relations because time expressions speak for themselves in
terms of reliability. That is, given a \KB triple $<$$e$, \texttt{rel}, $t$$>$ and its
aligned sentences,  the  finer-grained the time expression $t$ appears in the sentence,
the more likely the sentence  supports the existence of this triple.
For example, a sentence containing both \emph{Alphabet} and \emph{October-2-2015} is very likely to express the \texttt{inception-time} of \emph{Alphabet}, while a sentence containing both \emph{Alphabet} and \emph{2015} could instead talk  about many events, e.g.,  releasing financial report of 2015, hiring a new CEO, etc.
Using this heuristics, we can split the dataset into
3 subsets according to different granularities of the time expressions involved, indicating different levels of reliability.
Our criteria for determining the reliability are as follows.
Instances with full date expressions, i.e., \texttt{Year-Month-Day}, can be seen as the most reliable data, while those with
partial date expressions, e.g., \texttt{Month-Year} and \texttt{Year-Only}, are considered as less
reliable.  Negative data are constructed  heuristically that any
\emph{entity-time} pairs in a sentence without corresponding triples in Wikidata are treated as negative data.
During training, we can access  184,579 negative
 and  77,777 positive sentences, including 22,214 reliable, 
2,094 and 53,469 less reliable ones. The validation set and test set are randomly sampled from
the reliable (full-date) data for relatively fair evaluations and contains
2,776, 2,771 positive sentences and 5,143, 5,095 negative sentences, respectively.

\paragraph{\EntityRE} is a widely-used entity
relation extraction dataset, built 
by aligning triples
in Freebase to the New York Times (\NYT) corpus~\cite{riedel2010modeling}. 
It contains 52 relations, 136,947 positive and 385,664 negative sentences for training, and 6,444 positive and 166,004 negative sentences  for testing.
Unlike \TimeRE, this dataset does not contain any prior knowledge about the data quality.
Since the sentence level annotations in 
 \EntityRE are too noisy to serve as gold standard,  we only evaluate bag-level models on \EntityRE, a standard practice in previous works~\cite{surdeanu2012multi,zeng2015distant,lin2016neural}.

\subsection{Experimental Setup}
\paragraph{Hyper-parameters}
We use 200 convolution kernels with widow size 3. During training, we use stochastic gradient descend (SGD) with batch size 20.  The learning rates for sentence-level  and bag-level models are 0.1 and 0.01, respectively.
%
%

Sentence level experiments are performed on \TimeRE, 
using 
100-d word embeddings pre-trained using GloVe~\cite{pennington2014glove} on Wikipedia and Gigaword~\cite{parker2011english}, 
and 20-d vectors for distance embeddings. Each of the three subsets of \TimeRE is added after the previous phase has run for 15 epochs. The trace regularization weights are $\beta_1=0.01$, $\beta_2=-0.01$ and $\beta_3=-0.1$, respectively, from the reliable to the most unreliable, with the ratio of $\beta_3$ and $\beta_2$ fixed to 10 or 5 when tuning.

Bag level experiments are performed on both  \TimeRE and  \EntityRE. For \TimeRE, we use the same parameters as above.
For \EntityRE, we use 50-d word embeddings pre-trained on the \NYT corpus using word2vec~\cite{mikolov2013distributed}, 
and 5-d vectors for distance embedding.
For both datasets,  $\alpha$ and $\beta$ in Eq.~\ref{general_loss} are initialized to 1 and 0.1, respectively. We tried various decay rates, \{0.95, 0.9, 0.8\}, and steps, \{3, 5, 8\}. We found that using a decay rate of 0.9 with step of 5 gives best performance in most cases.

\paragraph{Evaluation Metric}
The performance is reported using the precision-recall (\PR) curve,
which is a standard evaluation metric in relation extraction.
Specifically, the extraction results are first ranked decreasingly by their confidence scores, then the precision and recall are calculated by setting the threshold to be the score of each extraction result one by one.

\paragraph{Naming Conventions}
We evaluate our approach under a wide range of settings for
 sentence level
(\texttt{sent\_}) and bag level (\texttt{bag\_}) models:
(1) \texttt{\_mix}:  trained on all three subsets of \TimeRE mixed together;
(2) \texttt{\_reliable}:  trained using the reliable subset of \TimeRE only;
(3) \texttt{\_PR}:  trained with prior knowledge of annotation quality, i.e., starting from the reliable data and then adding the unreliable data;
(4) \texttt{\_TM}: trained with dynamic transition matrix;
(5) \texttt{\_GTM}:  trained with a global transition matrix.
%
%
In bag level, we also investigate the performance   
of average aggregation (\texttt{\_avg})
and attention aggregation (\texttt{\_att}).

\section{Experimental Results \label{sec:evaluation}}

\subsection{Performance on \TimeRE} \label{sec:results_in_TimeRE}
\begin{figure}[t!]
\setlength{\belowcaptionskip}{-10pt}
\begin{center}
\includegraphics[width=0.45\textwidth]{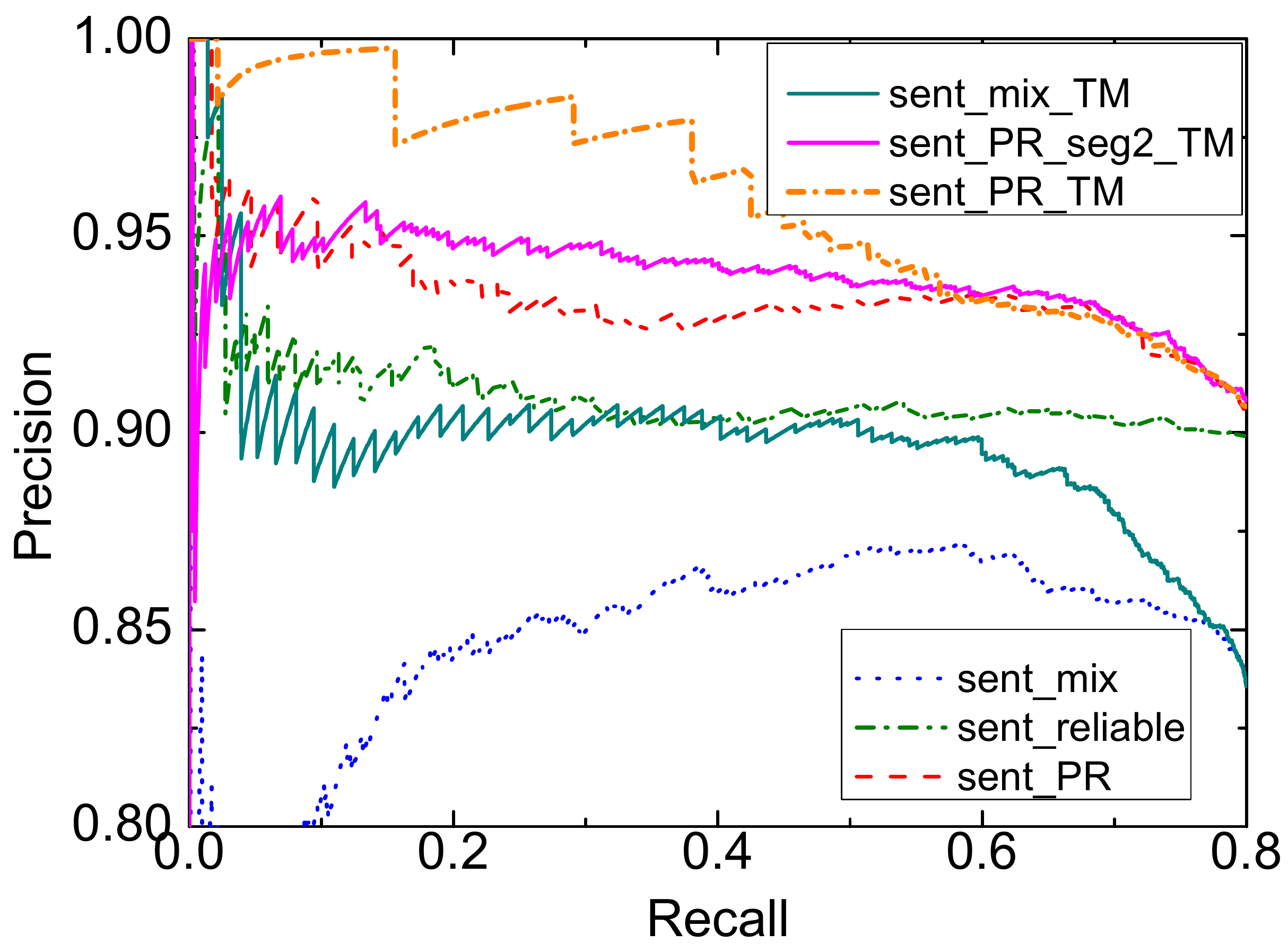}
\caption{Sentence Level Results on \TimeRE}
\label{fig: sent_luo}
\end{center}
\end{figure}

\paragraph{Sentence Level Models}
The results of sentence level models on \TimeRE are shown in Figure \ref{fig:
sent_luo}. We can see that mixing all subsets together (\texttt{sent\_mix})
gives the worst performance,  significantly
worse than using the reliable subset only (\texttt{sent\_reliable}). This
suggests the noisy nature of the training data obtained through \DS and
properly dealing with the noise is the key for \DS for a wider range of applications. 
When getting help from our dynamic transition matrix, 
the model (\texttt{sent\_mix\_TM}) significantly improves \texttt{sent\_mix},
delivering the same level of performance as \texttt{sent\_reliable} in most
cases. This suggests that our transition matrix can help to mitigate the bad influence of noisy training instances.

Now let us consider the \texttt{PR} scenario where one can build a curriculum by first training on the
reliable subset, then gradually moving to both reliable and less
reliable data. We can see that, this simple curriculum learning based model
(\texttt{sent\_PR}) further outperforms \texttt{sent\_reliable} significantly,
indicating that the curriculum learning framework not only reduces the effect
of noise, but also helps the model learn from noisy data. When applying the
transition matrix approach into this curriculum learning framework using one reliable
subset and one unreliable subset generated by mixing our two less reliable subsets, our model (\texttt{sent\_PR\_seg2\_TM})
further improves \texttt{sent\_PR} by  
utilizing the dynamic transition matrix to model the noise.
It is not surprising that when we use all three subsets separately,
our model (\texttt{sent\_PR\_TM}) significantly outperforms all
other models by a large margin.

\begin{figure*}[t!]
\setlength{\belowcaptionskip}{-10pt}
\centering
\subfigure[Attention Aggregation]{
\includegraphics[width=0.45\textwidth]{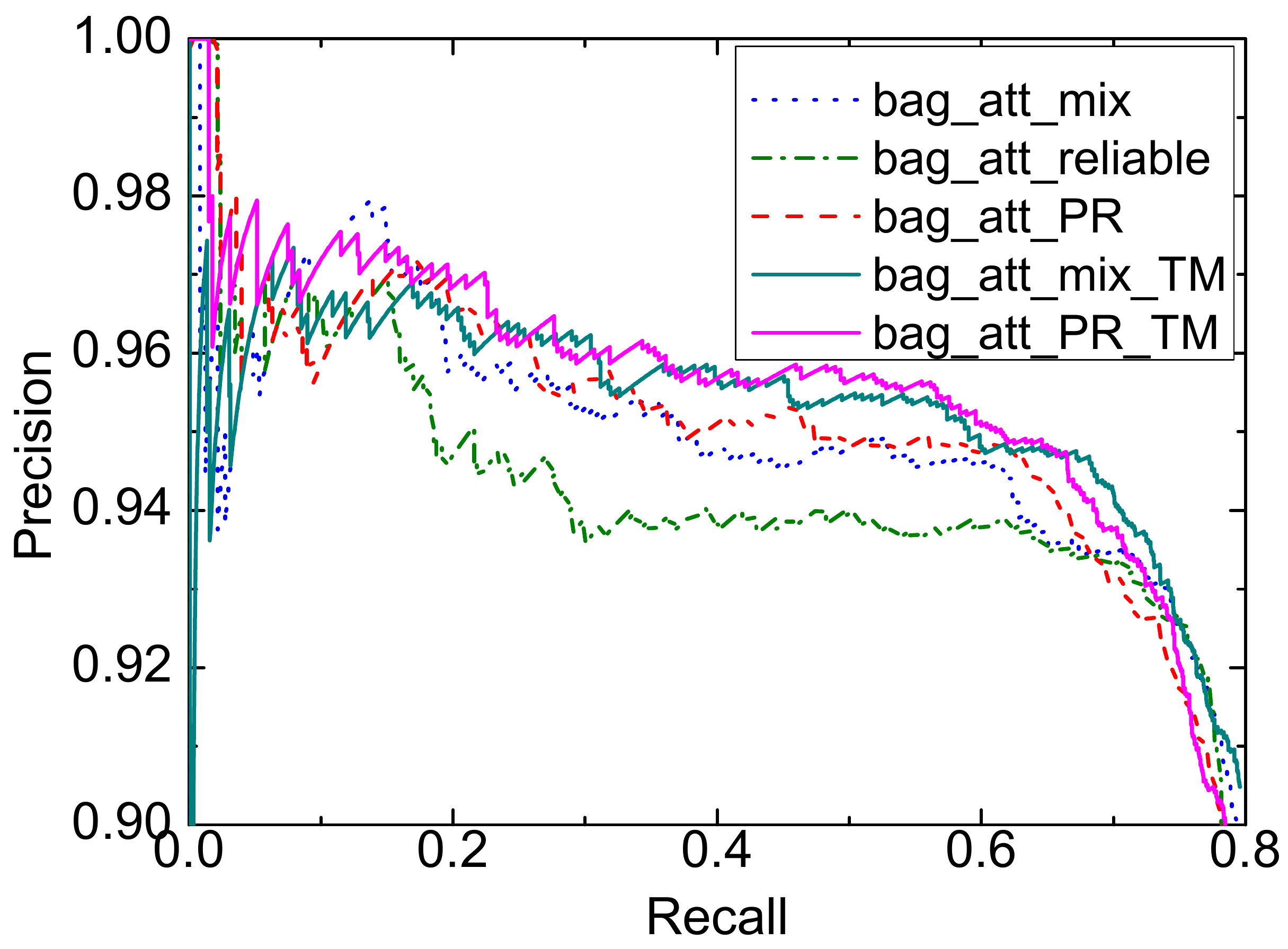}
\label{fig: bag_att_luo}
}
\subfigure[Average Aggregation]{
\includegraphics[width=0.45\textwidth]{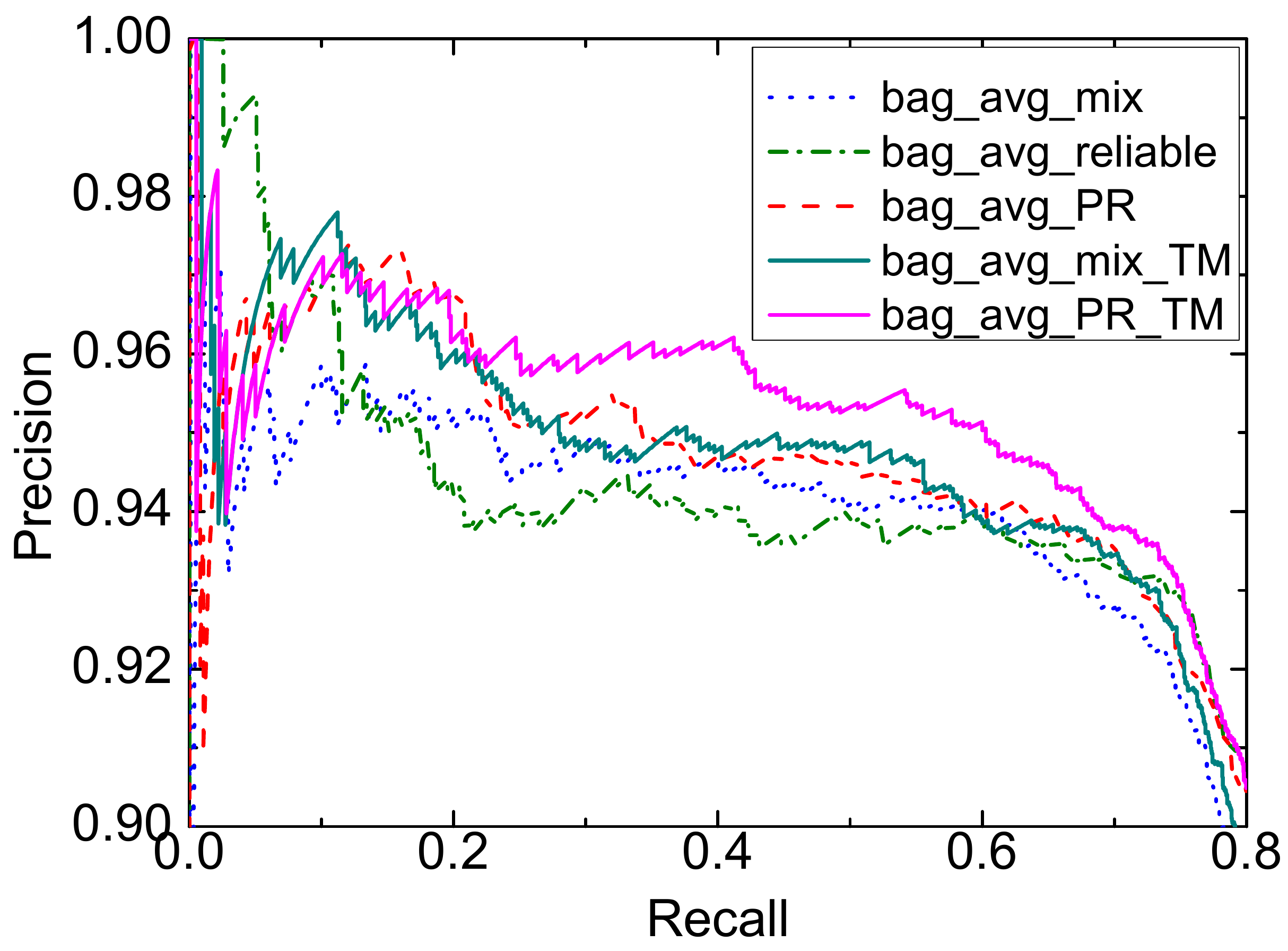}
\label{fig: bag_avg_luo}
}
\caption{Bag Level Results on \TimeRE}
\label{fig: results_on_luo}
\end{figure*}

\paragraph{Bag Level Models}
In this setting, we first look at the performance of the bag level models with attention aggregation. The results are shown in Figure~\ref{fig: bag_att_luo}.
Consider the comparison between the  model trained on the reliable subset only (\texttt{bag\_att\_reliable}) and the one trained on the mixed dataset (\texttt{bag\_att\_mix}).
In contrast to the sentence level, \texttt{bag\_att\_mix} outperforms \texttt{bag\_att\_reliable} by a large margin, because \texttt{bag\_att\_mix} has taken the \textit{at-least-one assumption} into consideration through the attention aggregation mechanism (Eq.~\ref{att_sum}), which can be seen as a denoising step within the bag.
This may also be the reason that when we introduce either our dynamic transition matrix (\texttt{bag\_att\_mix\_TM})  or the curriculum of using prior knowledge of data quality (\texttt{bag\_att\_PR}) into the bag level models, the improvement regarding \texttt{bag\_att\_mix}  is not as significant as in the sentence level.

However, when we apply our dynamic transition matrix 
into the curriculum built upon prior knowledge of data quality (\texttt{bag\_att\_PR\_TM}), the performance gets further improved. This happens especially in the high precision part compared to \texttt{bag\_att\_PR}.
We also note that the bag level's \textit{at-least-one assumption} does not always hold, and there are still false negative and false positive problems. Therefore, using our transition matrix approach with or without prior knowledge of data quality, i.e., \texttt{bag\_att\_mix\_TM}  and \texttt{bag\_att\_PR\_TM}, both improve the performance, and \texttt{bag\_att\_PR\_TM} performs slightly better.


The results of bag level models with average aggregation are shown in Figure \ref{fig: bag_avg_luo}, where the relative ranking of various settings is similar to those with attention aggregation.
A notable difference is that both \texttt{bag\_avg\_PR} and \texttt{bag\_avg\_mix\_TM} improve \texttt{bag\_avg\_mix} by a larger margin compared to that in the attention aggregation setting. The reason may be that the average aggregation mechanism is not as good as the attention aggregation in denoising within the bag, which leaves more space for our transition matrix approach or curriculum learning with prior knowledge to improve.
Also note that \texttt{bag\_avg\_reliable} performs best in the very-low-recall region but worst in general.
This is because that it ranks higher the sentences expressing either \texttt{birth-date} or \texttt{death-date}, the simplest but the most common relations in the dataset, but fails to learn other relations with limited or noisy training instances, given its relatively simple aggregation strategy.

\paragraph{Global v.s. Dynamic Transition Matrix}
\begin{figure}[t!]
\begin{center}
\includegraphics[width=0.45\textwidth]{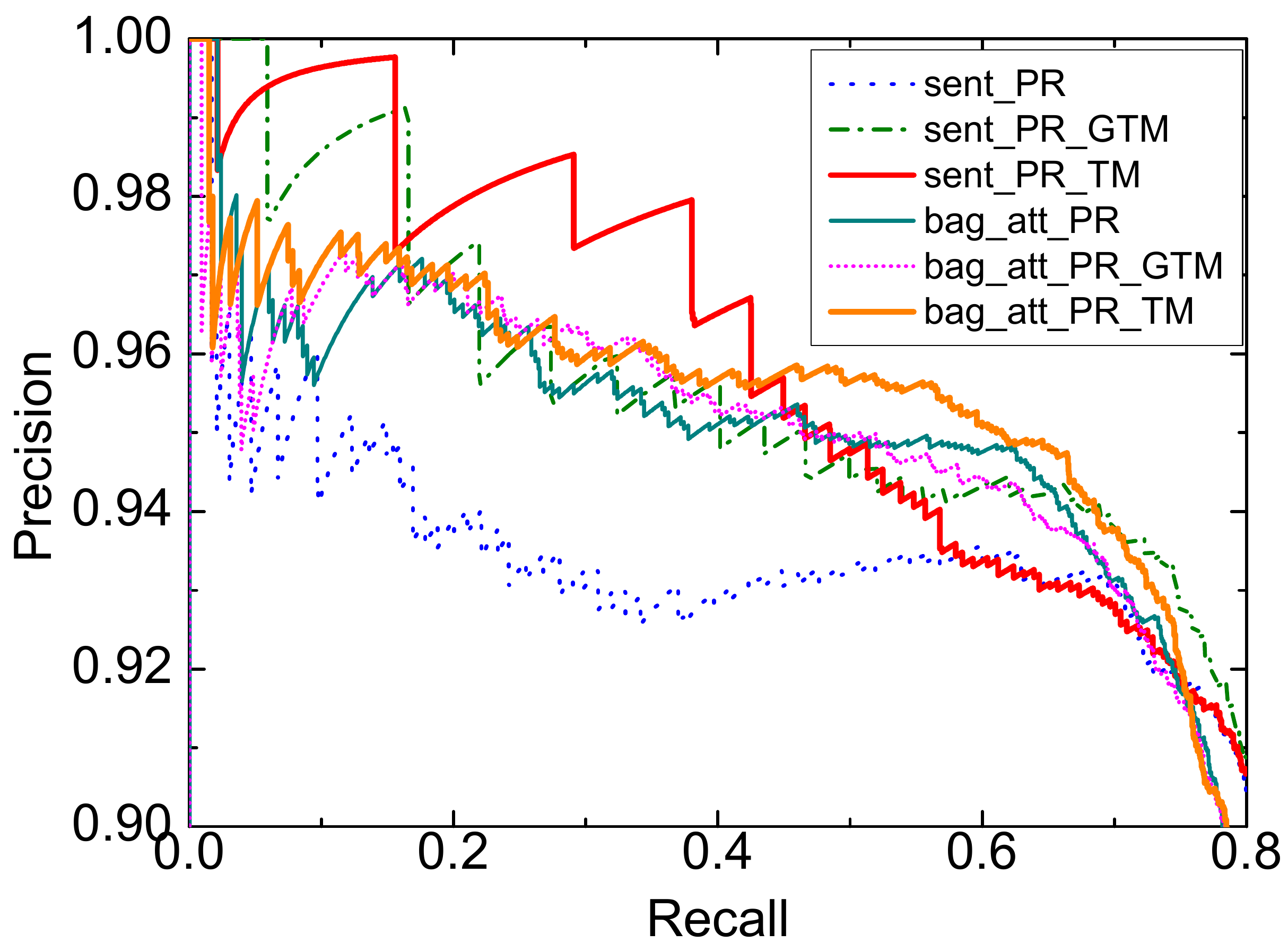}
\caption{Global TM v.s. Dynamic TM}
\label{fig: cmp_single_dynamic}
\end{center}
\end{figure}
We also compare our dynamic transition matrix method with the global transition matrix method,
which maintains only one transition matrix for all training instances. 
Specifically, instead of dynamically generating a transition matrix for each datum, we first initialize an identity matrix $\mathbf{T}'\in\mathbb{R}^{|\mathbb{C}|\times |\mathbb{C}|}$, where $|\mathbb{C}|$ is the number of relations (including \texttt{no-relation}).
Then the global transition matrix $\mathbf{T}$ is built by applying \emph{softmax} to each row of $\mathbf{T}'$ so that $\sum_j{\mathbf{T}_{ij}}=1$:
\begin{equation}
\label{shared_mat}
T_{ij} = \frac{e^{T'_{ij}}}{\sum_{j=1}^{|\mathbb{C}|}{e^{T'_{ij}}}}
\end{equation}
where $T_{ij}$ and $T'_{ij}$ are the elements in the $i^{th}$ row and $j^{th}$ column of $\mathbf{T}$ and $\mathbf{T}'$. The element values of matrix $\mathbf{T}'$ are also updated via backpropagation during training.
As shown in Figure~\ref{fig: cmp_single_dynamic},
using one global transition matrix (\texttt{\_GTM}) is also beneficial and improves both the sentence level (\texttt{sent\_PR}) and bag level  (\texttt{bag\_att\_PR}) models. However, since the global transition matrix only 
captures the global noise pattern, it fails to characterize individuals with subtle differences, 
resulting in a performance drop compared to the dynamic one (\texttt{\_TM}).

\paragraph{Case Study}
We find our transition matrix method tends to obtain more significant improvement on noisier relations. For example, \textit{time\_of\_spacecraft\_landing} is noisier than \textit{time\_of\_spacecraft\_launch} since compared to the launching of a spacecraft, there are fewer sentences containing the landing time of a spacecraft that talks directly about the landing. Instead, many of these sentences tend to talk about the activities of the crew. Our \texttt{sent\_PR\_TM} model improves the F1 of \textit{time\_of\_spacecraft\_landing} and \textit{time\_of\_spacecraft\_launch} over \texttt{sent\_PR} by 9.09\% and 2.78\%, respectively. 
The transition matrix makes more significant improvement on \textit{time\_of\_spacecraft\_landing} since there are more noisy sentences for our method to handle, which results in more significant improvement on the quality of the training data.

\subsection{Performance on \EntityRE}
\begin{figure}[t!]
\includegraphics[width=0.45\textwidth]{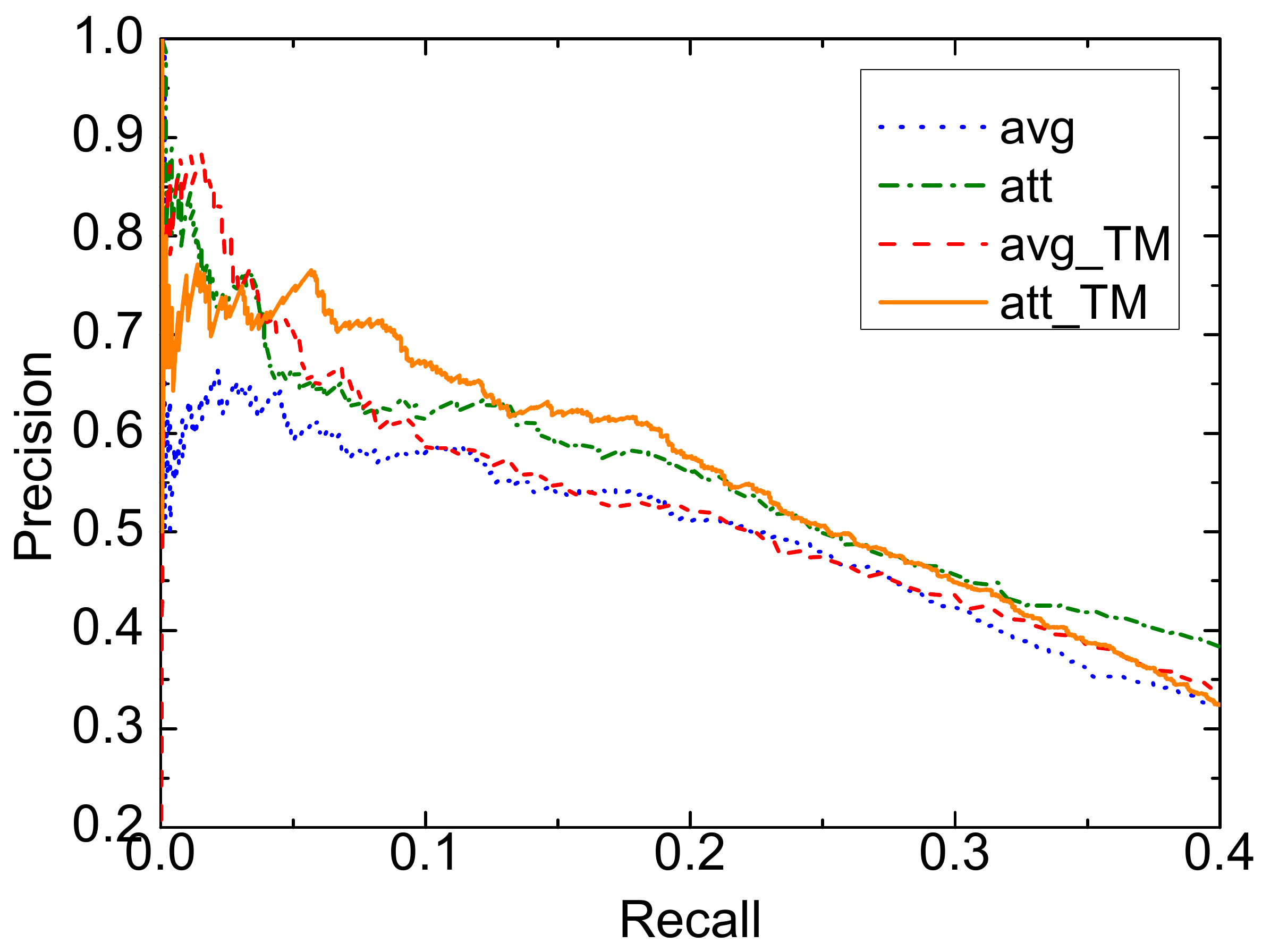}
\caption{Results on \EntityRE}
\label{fig: Riedel_res}
\end{figure}

We evaluate our bag level models on \EntityRE.
As shown in Figure~\ref{fig: Riedel_res}, it is not surprising that the basic model with attention aggregation (\texttt{att}) significantly outperforms the average one (\texttt{avg}), where \texttt{att} in our bag embedding is similar in spirit to \cite{lin2016neural},
which has reported the-state-of-the-art performance on \EntityRE.
When injected with our transition matrix approach,  both \texttt{att\_TM} and \texttt{avg\_TM} clearly outperform their basic versions.

Similar to the situations in \TimeRE, since \texttt{att} has taken the \textit{at-least-one assumption} into account through its attention-based bag embedding mechanism, thus the improvement made by \texttt{att\_TM} is not as large as by \texttt{avg\_TM}.

\begin{table}
\centering
\small{
\begin{tabular}{|c|c|c|c|c|}
\hline
\textbf{Method}							& \textbf{P@R\_10} 		& \textbf{P@R\_20} 			& \textbf{P@R\_30} \\
\hline
\textit{Mintz} 							&39.88	&28.55	&16.81 	\\
\hline
\textit{MultiR} 						&60.94	&36.41	&- 	\\
\hline
\textit{MIML} 							&60.75	&33.82	&- 	\\
\hline
\textit{avg} 								&58.04	&51.25	&42.45 	\\
\hline
\textit{avg\_TM} 						&58.56	&52.35	&43.59 	\\
\hline
\textit{att} 								&61.51	&56.36	&\textbf{45.63} 	\\
\hline
\textit{att\_TM} 						&\textbf{67.24}	&\textbf{57.61}	&44.90 	\\
\hline
\end{tabular}
}
\caption{Comparison with feature-based methods. P@R\_10/20/30 refers to the precision when recall equals 10\%, 20\% and 30\%.}
\label{feature-based}
\end{table}

We also include the comparison with three feature-based methods: \texttt{Mintz}~\cite{mintz2009distant} is a multiclass logistic regression model; \texttt{MultiR}~\cite{hoffmann2011knowledge} is a probabilistic graphical model that can handle overlapping relations; \texttt{MIML}~\cite{surdeanu2012multi} is also a probabilistic graphical model but operates in the multi-instance multi-label paradigm. As shown in Table~\ref{feature-based}, although traditional feature-based methods have reasonable results in the low recall region, their performances drop quickly as the recall goes up,
and \texttt{MultiR} and \texttt{MIML} did not even reach the 30\% recall.
This indicates that, while human-designed featurs can effectively capture certain relation patterns, their coverage is relatively low.
On the other hand, neural network models have more stable performance across different recalls, 
and \texttt{att\_TM} performs generally better than other models, 
indicating again the effectiveness of our transition matrix method.

\section{Related Work}

In addition to relation extraction, 
distant supervision (\DS) is shown to be effective in generating training data for various NLP tasks, e.g., tweet sentiment
classification~\cite{go2009twitter}, tweet named entity classifying~\cite{ritter2011named}, etc.
However, these early applications of \DS do not well address the issue of data noise.


In relation extraction (\RE), recent works  have been proposed to reduce the influence of wrongly labeled data.
The work presented by \cite{takamatsu2012reducing} removes potential noisy sentences by identifying bad syntactic
patterns at the pre-processing stage. \cite{xu2013filling} use pseudo-relevance feedback to find
possible false negative data. 
\cite{riedel2010modeling} make the \emph{at-least-one assumption} 
and 
propose to alleviate the noise problem by considering \RE  as a multi-instance classification problem.
Following this assumption, people further improves the original paradigm using probabilistic graphic models~\cite{hoffmann2011knowledge,surdeanu2012multi}, and neural network methods \cite{zeng2015distant}. 
Recently, \cite{lin2016neural} propose to use attention mechanism to reduce the noise within a sentence bag. 
Instead of characterizing the noise, these approaches only aim to alleviate the effect of noise. 
%

The \emph{at-least-one assumption} is often too strong in practice, and there are still chances that the
sentence bag may be false positive or false negative. Thus it is important to model the noise pattern 
to guide the learning procedure.
\cite{ritter2013modeling} and \cite{min2013distant} try to 
employ a set of latent variables to represent the true relation. Our approach differs
from them in two aspects. 
We target  noise modeling in neutral networks while they target probabilistic graphic models. 
We further advance their models by providing the capability to model the fine-grained transition from the true relation to the observed, and 
the flexibility to combine indirect guidance.
Outside of NLP, various methods have been proposed in computer vision  to 
 model the data noise using neural networks.
\cite{sukhbaatar2014training}  utilize a global transition matrix with weight decay to transform the true label distribution to 
the observed.  
\cite{reed2014training}  use a hidden layer to represent the true label distribution but try to force it to predict both the noisy label and the input. \cite{chen2015webly,xiao2015learning} first estimate the transition matrix on a clean dataset and apply to the noisy data. 
Our model shares similar spirit with \cite{misra2016seeing} in that we all dynamically generate a transition matrix for each
training instance, but, instead of using vanilla SGD, we train our model with a novel curriculum learning training framework with trace regularization to control the behavior of transition matrix.
In NLP, the only work in neural-network-based noise modeling is to use one single global transition matrix to model the noise introduced by
cross-lingual projection of training data~\cite{fang2016learning}.
 Our work advances them through generating a transition matrix dynamically for each instance, to avoid  using one single component to characterize both reliable and unreliable data. 

\vspace{-.4em}\section{Conclusions}
In this paper, we investigate the noise problem inherent in  the \DS-style training data.
 We argue that the data speak for themselves by providing useful clues to reveal their noise patterns.
We thus propose a novel transition matrix based method to dynamically characterize the noise underlying such training data in a unified framework along  the original prediction objective.
One of our key innovations is to exploit a curriculum learning based training method to gradually  learn to model  the underlying noise pattern without direct guidance, and to provide the flexibility to exploit any prior knowledge of the data quality to further improve the effectiveness of the transition matrix.
We evaluate our approach in two learning settings of the distantly supervised relation extraction. The experimental results show
that the proposed method can better characterize the underlying noise and consistently outperform start-of-the-art extraction models under various scenarios.

\section*{Acknowledgement}
This work is supported by the National High Technology R\&D Program of China (2015AA015403); 
the National Natural Science Foundation of China (61672057, 61672058);
KLSTSPI Key Lab. of Intelligent Press Media Technology;
the UK Engineering and Physical Sciences Research Council under
grants EP/M01567X/1 (SANDeRs) and EP/M015793/1 (DIVIDEND);
and the Royal Society International Collaboration Grant (IE161012).


\bibliography{acl2017}

\begin{thebibliography}{}
\expandafter\ifx\csname natexlab\endcsname\relax\def\natexlab#1{#1}\fi

\bibitem[{Bengio et~al.(2009)Bengio, Louradour, Collobert, and
  Weston}]{bengio2009curriculum}
Yoshua Bengio, J{\'e}r{\^o}me Louradour, Ronan Collobert, and Jason Weston.
  2009.
\newblock Curriculum learning.
\newblock In {\em ICML\/}. ACM, pages 41--48.

\bibitem[{Chen and Gupta(2015)}]{chen2015webly}
Xinlei Chen and Abhinav Gupta. 2015.
\newblock Webly supervised learning of convolutional networks.
\newblock In {\em ICCV\/}. pages 1431--1439.

\bibitem[{Fang and Cohn(2016)}]{fang2016learning}
Meng Fang and Trevor Cohn. 2016.
\newblock Learning when to trust distant supervision: An application to
  low-resource pos tagging using cross-lingual projection.
\newblock In {\em CONLL\/}. pages 178--186.

\bibitem[{Go et~al.(2009)Go, Bhayani, and Huang}]{go2009twitter}
Alec Go, Richa Bhayani, and Lei Huang. 2009.
\newblock Twitter sentiment classification using distant supervision.
\newblock {\em CS224N Project Report, Stanford\/} 1(12).

\bibitem[{Hoffmann et~al.(2011)Hoffmann, Zhang, Ling, Zettlemoyer, and
  Weld}]{hoffmann2011knowledge}
Raphael Hoffmann, Congle Zhang, Xiao Ling, Luke Zettlemoyer, and Daniel~S Weld.
  2011.
\newblock Knowledge-based weak supervision for information extraction of
  overlapping relations.
\newblock In {\em Proceedings of ACL\/}. pages 541--550.

\bibitem[{Lin et~al.(2016)Lin, Shen, Liu, Luan, and Sun}]{lin2016neural}
Yankai Lin, Shiqi Shen, Zhiyuan Liu, Huanbo Luan, and Maosong Sun. 2016.
\newblock Neural relation extraction with selective attention over instances.
\newblock In {\em ACL\/}. volume~1, pages 2124--2133.

\bibitem[{Mikolov et~al.(2013)Mikolov, Sutskever, Chen, Corrado, and
  Dean}]{mikolov2013distributed}
Tomas Mikolov, Ilya Sutskever, Kai Chen, Greg~S Corrado, and Jeff Dean. 2013.
\newblock Distributed representations of words and phrases and their
  compositionality.
\newblock In {\em NIPS\/}. pages 3111--3119.

\bibitem[{Min et~al.(2013)Min, Grishman, Wan, Wang, and
  Gondek}]{min2013distant}
Bonan Min, Ralph Grishman, Li~Wan, Chang Wang, and David Gondek. 2013.
\newblock Distant supervision for relation extraction with an incomplete
  knowledge base.
\newblock In {\em HLT-NAACL\/}. pages 777--782.

\bibitem[{Mintz et~al.(2009)Mintz, Bills, Snow, and
  Jurafsky}]{mintz2009distant}
Mike Mintz, Steven Bills, Rion Snow, and Dan Jurafsky. 2009.
\newblock Distant supervision for relation extraction without labeled data.
\newblock In {\em ACL\/}. pages 1003--1011.

\bibitem[{Misra et~al.(2016)Misra, Lawrence~Zitnick, Mitchell, and
  Girshick}]{misra2016seeing}
Ishan Misra, C~Lawrence~Zitnick, Margaret Mitchell, and Ross Girshick. 2016.
\newblock Seeing through the human reporting bias: Visual classifiers from
  noisy human-centric labels.
\newblock In {\em CVPR\/}. pages 2930--2939.

\bibitem[{Parker et~al.(2011)Parker, Graff, Kong, Chen, and
  Maeda}]{parker2011english}
Robert Parker, David Graff, Junbo Kong, Ke~Chen, and Kazuaki Maeda. 2011.
\newblock English gigaword fifth edition, linguistic data consortium.
\newblock Technical report, Linguistic Data Consortium, Philadelphia.

\bibitem[{Pennington et~al.(2014)Pennington, Socher, and
  Manning}]{pennington2014glove}
Jeffrey Pennington, Richard Socher, and Christopher~D Manning. 2014.
\newblock Glove: Global vectors for word representation.
\newblock In {\em EMNLP\/}. volume~14, pages 1532--1543.

\bibitem[{Reed et~al.(2014)Reed, Lee, Anguelov, Szegedy, Erhan, and
  Rabinovich}]{reed2014training}
Scott Reed, Honglak Lee, Dragomir Anguelov, Christian Szegedy, Dumitru Erhan,
  and Andrew Rabinovich. 2014.
\newblock Training deep neural networks on noisy labels with bootstrapping.
\newblock {\em arXiv preprint arXiv:1412.6596\/} .

\bibitem[{Riedel et~al.(2010)Riedel, Yao, and McCallum}]{riedel2010modeling}
Sebastian Riedel, Limin Yao, and Andrew McCallum. 2010.
\newblock Modeling relations and their mentions without labeled text.
\newblock In {\em Joint European Conference on Machine Learning and Knowledge
  Discovery in Databases\/}. Springer, pages 148--163.

\bibitem[{Ritter et~al.(2011)Ritter, Ritter, Clark, Etzioni
  et~al.}]{ritter2011named}
Alan Ritter, Alan Ritter, Sam Clark, Oren Etzioni, et~al. 2011.
\newblock Named entity recognition in tweets: an experimental study.
\newblock In {\em EMNLP\/}. Association for Computational Linguistics, pages
  1524--1534.

\bibitem[{Ritter et~al.(2013)Ritter, Zettlemoyer, Mausam, and
  Etzioni}]{ritter2013modeling}
Alan Ritter, Luke Zettlemoyer, Mausam, and Oren Etzioni. 2013.
\newblock Modeling missing data in distant supervision for information
  extraction.
\newblock {\em TACL\/} 1:367--378.

\bibitem[{Sukhbaatar et~al.(2015)Sukhbaatar, Bruna, Paluri, Bourdev, and
  Fergus}]{sukhbaatar2014training}
Sainbayar Sukhbaatar, Joan Bruna, Manohar Paluri, Lubomir Bourdev, and Rob
  Fergus. 2015.
\newblock Training convolutional networks with noisy labels.
\newblock In {\em ICLR\/}.

\bibitem[{Surdeanu et~al.(2012)Surdeanu, Tibshirani, Nallapati, and
  Manning}]{surdeanu2012multi}
Mihai Surdeanu, Julie Tibshirani, Ramesh Nallapati, and Christopher~D Manning.
  2012.
\newblock Multi-instance multi-label learning for relation extraction.
\newblock In {\em EMNLP-CoNLL\/}. pages 455--465.

\bibitem[{Takamatsu et~al.(2012)Takamatsu, Sato, and
  Nakagawa}]{takamatsu2012reducing}
Shingo Takamatsu, Issei Sato, and Hiroshi Nakagawa. 2012.
\newblock Reducing wrong labels in distant supervision for relation extraction.
\newblock In {\em ACL\/}. pages 721--729.

\bibitem[{Vrande{\v{c}}i{\'c} and Kr{\"o}tzsch(2014)}]{vrandevcic2014wikidata}
Denny Vrande{\v{c}}i{\'c} and Markus Kr{\"o}tzsch. 2014.
\newblock Wikidata: a free collaborative knowledgebase.
\newblock {\em Communications of the ACM\/} 57(10):78--85.

\bibitem[{Xiao et~al.(2015)Xiao, Xia, Yang, Huang, and Wang}]{xiao2015learning}
Tong Xiao, Tian Xia, Yi~Yang, Chang Huang, and Xiaogang Wang. 2015.
\newblock Learning from massive noisy labeled data for image classification.
\newblock In {\em CVPR\/}. pages 2691--2699.

\bibitem[{Xu et~al.(2013)Xu, Hoffmann, Zhao, and Grishman}]{xu2013filling}
Wei Xu, Raphael Hoffmann, Le~Zhao, and Ralph Grishman. 2013.
\newblock Filling knowledge base gaps for distant supervision of relation
  extraction.
\newblock In {\em ACL\/}. pages 665--670.

\bibitem[{Zeng et~al.(2015)Zeng, Liu, Chen, and Zhao}]{zeng2015distant}
Daojian Zeng, Kang Liu, Yubo Chen, and Jun Zhao. 2015.
\newblock Distant supervision for relation extraction via piecewise
  convolutional neural networks.
\newblock In {\em EMNLP\/}. pages 1753--1762.

\end{thebibliography}
\bibliographystyle{acl_natbib}

\end{document}